\documentclass[acmtog]{acmart}
\acmSubmissionID{251}

\usepackage{booktabs} 
\usepackage{bm}
\usepackage{enumitem}

\usepackage[ruled]{algorithm2e} 

\SetAlFnt{\small}
\SetAlCapFnt{\small}
\SetAlCapNameFnt{\small}
\SetAlCapHSkip{0pt}

\setcopyright{rightsretained}

\acmDOI{nnnnnnn.nnnnnn}

\newcommand{\EPos}{\bm{c}}
\newcommand{\EDir}{\bm{\omega}_o}
\newcommand{\LDir}{\bm{\omega}_i}
\newcommand{\LPos}{\bm{l}}
\newcommand{\PosX}{\bm{x}}
\newcommand{\PosXPr}{\bm{x}^\prime}
\newcommand{\posx}{t}
\newcommand{\posxPr}{t^\prime}
\newcommand{\SNormal}{\bm{n}}
\newcommand{\SReflect}{\bm{R}}
\newcommand{\Radiance}{L}
\newcommand{\EN}{N}

\newcommand{\Extinct}{\sigma}

\newcommand{\Trans}{\tau}
\newcommand{\Import}{a}

\newcommand{\cnrf}{Neural Reflectance Fields}
\newcommand{\nrf}{neural reflectance field}
\newcommand{\nrfs}{neural reflectance fields}

\newcommand{\SHouse}{\textsc{House}}

\newcommand{\SDragon}{\textsc{Dragon}}
\newcommand{\SPlane}{\textsc{Plane}}
\newcommand{\SShop}{\textsc{Shop}}
\newcommand{\SAMan}{\textsc{Superhero}}

\hypersetup{draft}

\begin{document}

\title{\cnrf{} for Appearance Acquisition}


\newcommand{\Comment}[1]{}

\author{Sai Bi}
\authornote{Both authors contributed equally to this research.}
\affiliation{
  \institution{UC San Diego}
}
\email{bisai@cs.ucsd.edu}
\author{Zexiang Xu}
\authornotemark[1]
\affiliation{
  \institution{UC San Diego, Adobe Research}
}
\email{zexu@adobe.com}

\author{Pratul Srinivasan}
\email{pratul@berkeley.edu}
\author{Ben Mildenhall}
\email{bmild@cs.berkeley.edu}
\affiliation{
  \institution{UC Berkeley}
}

\author{Kalyan Sunkavalli}
\email{sunkaval@adobe.com}
\author{Milo\v{s} Ha\v{s}an}
\email{mihasan@adobe.com}
\author{Yannick Hold-Geoffroy}
\email{holdgeof@adobe.com}
\affiliation{
  \institution{Adobe Research}
}

\author{David Kriegman}
\email{kriegman@cs.ucsd.edu}
\author{Ravi Ramamoorthi}
\email{ravir@cs.ucsd.edu}
\affiliation{
  \institution{UC San Diego}
}


\begin{abstract}
We present \emph{\nrfs{}}, a novel deep scene representation that encodes volume density, normal and reflectance properties at any 3D point in a scene using a fully-connected neural network. We combine this representation with a physically-based differentiable ray marching framework that can render images from a \nrf{} under any viewpoint and light. We demonstrate that neural reflectance fields can be estimated from images captured with a simple collocated camera-light setup, and accurately model the appearance of real-world scenes with complex geometry and reflectance. Once estimated, they can be used to render photo-realistic images under novel viewpoint and (non-collocated) lighting conditions and accurately reproduce challenging effects like specularities, shadows and occlusions. This allows us to perform high-quality view synthesis and relighting that is significantly better than previous methods. We also demonstrate that we can compose the estimated \nrf{} of a real scene with traditional scene models and render them using standard Monte Carlo rendering engines. Our work thus enables a complete pipeline from high-quality and practical appearance acquisition to 3D scene composition and rendering.
\end{abstract}

\Comment{

}

\begin{teaserfigure}
    \includegraphics[width=\linewidth]{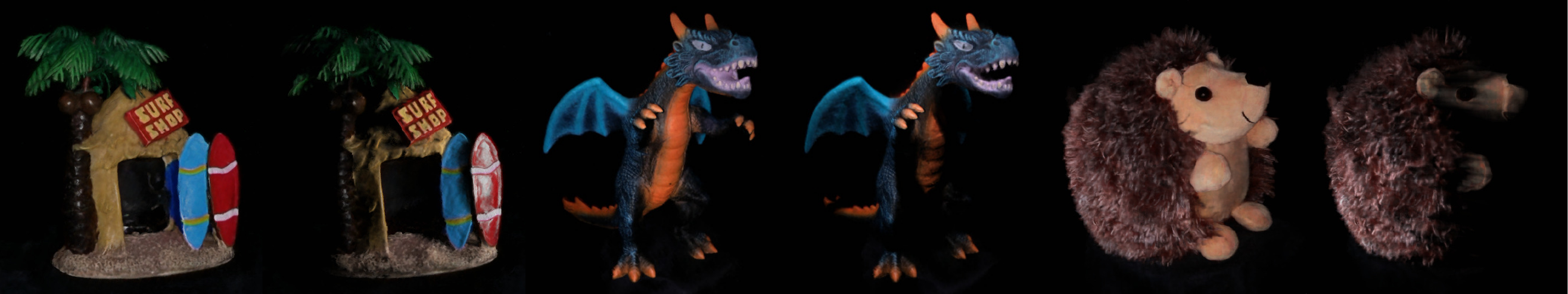}
    \caption{Renderings from our novel \nrf{} representation that is reconstructed from images of these scenes captured with a simple collocated camera-light setup. 
    Our \nrfs{} can model complex scene geometry and reflectance and render images from new viewpoints (odd images) and non-collocated lighting (even images) that were never captured. }
    \label{fig:teaser}
\end{teaserfigure}

\maketitle

\section{Introduction}
\label{sec:intro}
Modeling a real scene from captured images and 
reproducing its appearance under novel conditions is a central problem in computer graphics and vision.
This has traditionally been accomplished by using 3D reconstruction and inverse rendering methods to recover scene geometry and reflectance \cite{zhou2013multi,nam2018practical}.
However, this is an extremely challenging task, and even state-of-the-art methods generate inaccurate reconstructions that produce images with significant artifacts when rendered.

More recently, many approaches have been proposed that circumvent the problem of explicit reconstruction, and instead estimate a ``neural'' scene representations that can be combined with an appropriate differentiable rendering method to generate novel images (see ~\cite{tewari2020STAR} for a recent survey). 
One line of work in this space combines neural scene representations with classical ray marching---a volume rendering approach that is naturally differentiable---to achieve realistic rendering without requiring any pre-acquired 3D geometry \cite{lombardi2019neural,mildenhall2020nerf,sitzmann2019scene}.
However, these methods are mostly designed for view synthesis and do not model scene appearance as a function of reflectance or lighting. 
As a result, they do not allow for tasks such as relighting or scene editing.
While ray marching can be used with discrete volumes with explicit per-voxel BRDFs \cite{bi2020deep}
to enable both relighting and view synthesis,
an explicit discretized volume representation is highly restricted by its fixed resolution, 
and cannot reproduce high-frequency appearance details like fine textures and sharp boundaries.

\begin{figure}[t]
    \includegraphics[width=\linewidth]{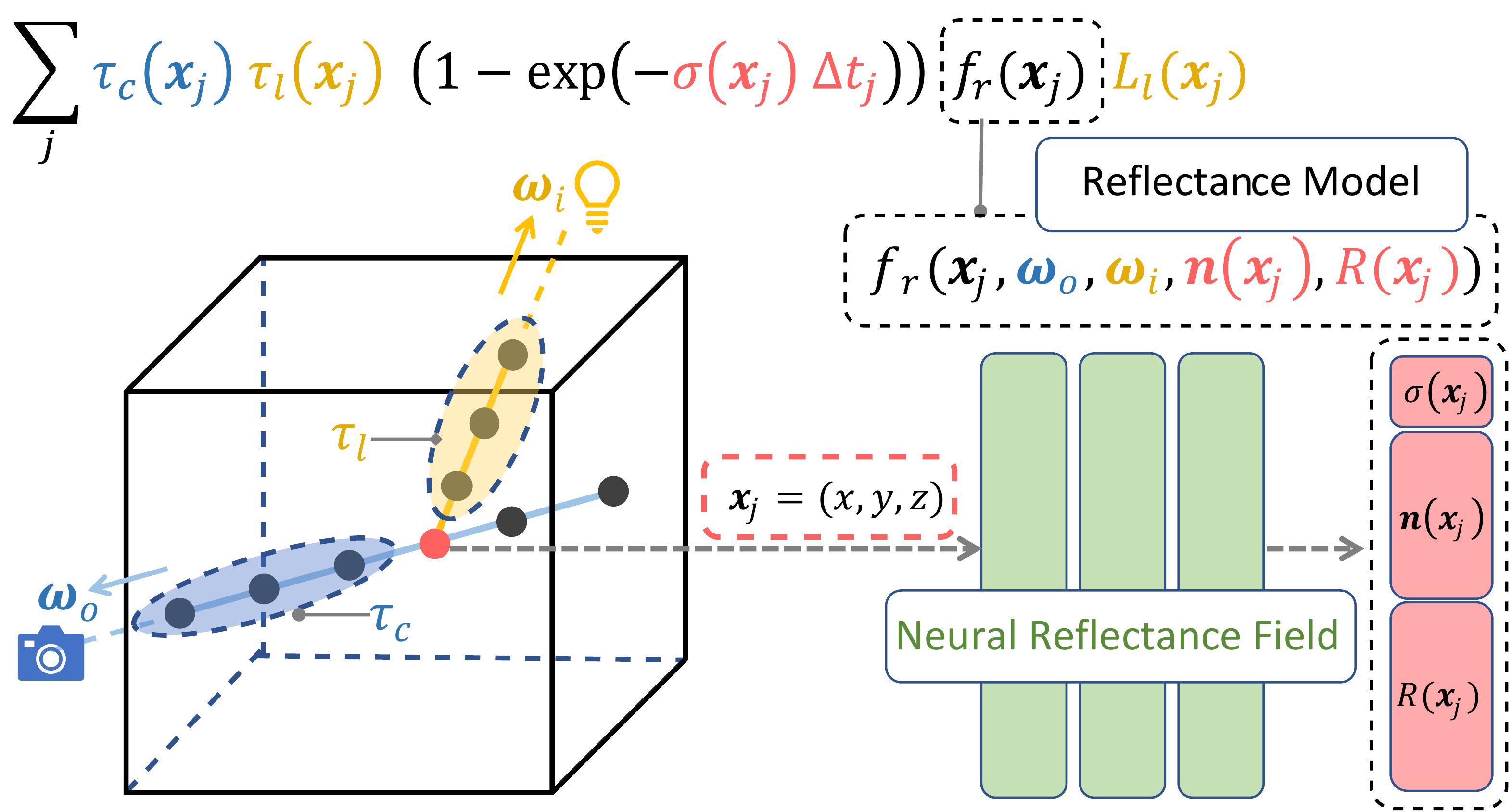}
    \caption{Overview of the neural reflectance field and ray marching. 
    We march a ray (blue) through each pixel from the camera and sample a sequence of shading points on the ray.
    The ray radiance under a point light source is computed from the volume density $\Extinct$, normal $\SNormal$ and reflectance properties $\SReflect$ via a differentiable ray marching equation (Eqn.~\ref{eqn:numrendering}) shown on top.
    At each point $\PosX_j$, shading is computed from the normal $\SNormal(\PosX_j)$, reflectance properties $\SReflect(\PosX_j)$ and corresponding light and view directions ($\LDir$ and $\EDir$) using a specified reflectance model $f_r$.
    Volume density $\Extinct(\PosX_j)$ acts like opacity ($\alpha=1-\exp(-\Extinct\Delta t)$) to attenuate the shading; it is also used to compute the view transmittance $\Trans_{\EPos}$ (Eqn.~\ref{eqn:numETrans}) along the camera ray (blue ellipsoid) and the light transmittance $\Trans_{\LPos}$ (Eqn.~\ref{eqn:numLTrans}) along an additional ray toward the light (yellow ellipsoid). 
    We propose to use a novel neural reflectance field, represented by an MLP, to regress the required rendering properties ($\Extinct$, $\SNormal$, $\SReflect$) from the 3D location $\PosX_j = (x,y,z)$ for ray marching.  }
    \label{fig:overview}
\end{figure}

In this work, we propose a novel scene representation that we refer to as \emph{\cnrf{}}. Unlike previous work that models scene color~\cite{lombardi2019neural} or radiance~\cite{mildenhall2020nerf}, \nrfs{} account for both scene geometry and \emph{reflectance}.
This allows us to combine \nrfs{} with a ray-marching framework 
(see Fig.~\ref{fig:overview}) to render images under arbitrary view and lighting.
Moreover, the whole pipeline is differentiable allowing us to pose the problem of appearance acquisition as one of \emph{optimizing} for a \nrf{} that, when rendered, will match the captured scene images.
Based on this, we capture multiple images around the scene with a cellphone camera and its built-in flash, similar to the acquisition in recent work on material acquisition~\cite{li2018materials,deschaintre2018single}, relighting~\cite{xu2018deep} and inverse rendering~\cite{nam2018practical}.
This practical setup yields unstructured multi-view images under collocated point illumination and captures complex high-frequency scene appearance.
As we illustrate in Fig.~\ref{fig:teaser}, \nrfs{} can be reconstructed from such ``simple'' inputs and allow for the photo-realistic rendering of  complex real scenes under novel viewpoints and lighting conditions (that are arbitrarily different from the captured collocated lighting).

\cnrf{} are a continuous function neural representation that implicitly models both scene geometry and reflectance.
We represent them using a deep multi-layer perceptron (MLP) that can regress reflectance properties, normals, and volume density at a given 3D scene point $(x, y, z)$.
This representation can be combined with a differentiable ray marching framework---based on classical physically based volume rendering \cite{kniss2003model,novak2018monte,max1995optical}.
In particular, we march rays from the viewpoint through each pixel, and along each marching ray sample points where we compute shading using the regressed normal and reflectance properties at sampled shading points. 
This shading is modulated with transmittance (computed from regressed volume density), 
and accumulated along the ray to compute radiance.

We utilize the transmittance not only along the camera ray but also along the light ray to model light effects like shadows for complex real scenes (see Fig.~\ref{fig:teaser}).
Naively computing the light transmittance requires marching a rays from all the points sampled along the camera ray to the light, making it intractable both for reconstruction and rendering.
Instead, we note that the collocated nature of our input data simplifies this for us because it only requires us to evaluate transmittance along the identical camera-light ray, thus allowing us to efficiently fit \nrfs{} to image data.
To further speed up re-rendering under arbitrary light and view positions, we pre-compute a light transmittance volume at adaptively sampled points, enabling efficient shadow rendering. 

Our entire rendering pipeline is general and can support any network that can map a 3D point to rendering parameters and any differentiable reflectance model.
For example, in Fig.~\ref{fig:teaser},~\ref{fig:more_results},~\ref{fig:face}, we demonstrate that we can accurately model the appearance of a diverse set of real scenes,
including scenes with intricate geometry, highly specular reflectance, furry objects, and human portraits.
These results are significantly better than the state-of-the-art mesh-based reconstruction method \cite{nam2018practical} and discrete volume-based representations \cite{bi2020deep} (see Fig.~\ref{fig:comp}).

Moreover, because our representation is designed to work with a physically-based volume renderer, it can in fact be naturally incorporated into modern rendering engines, like Mitsuba \cite{mitsuba}.
This allows us to compose \nrfs{} with traditional 3D models (with explicit meshes and BRDFs) and capture light transport interactions between these disparate scene elements (see Fig.~\ref{fig:mituba}).
This is something that has not been demonstrated by previous neural reconstruction methods, and in our opinion, represents an important step towards building neural capture and rendering approaches that can be incorporated into traditional 3D design workflows.


In summary, our main contributions are:
\begin{itemize}[noitemsep,topsep=0pt,wide, labelwidth=!, labelindent=0pt]
    \item A novel neural reflectance field representation that models both scene geometry and reflectance, 
    \item A physically-based ray marching scheme that can render \nrfs{} under any view and lighting,
    \item A method to reconstruct \nrfs{} from unstructured flash images, and
    \item Applications of this representation to tasks like view synthesis, relighting, and scene composition.
\end{itemize}

\Comment{
    Our reflectance-aware ray marching enables meaningful appearance reasoning in learning the neural representations;
this allows the networks to effectively regress scene reflectance can be used to render under any lighting and viewpoint.

    In our recent unpublished technical draft \cite{bi2020deep} 
    \footnote{The draft \shortcite{bi2020deep} is 
    currently under review. We anonymously include the preprint in our supplemental material to facilitate the reviewing of this submission.},  
    by also leveraging ray marching in a deep learning framework, 
    we explicitly reconstruct a scene as reflectance volumes with per-voxel BRDFs, 
    which enables both relighting and view synthesis.
    However, such explicit volume representation is highly restricted by its fixed resolution, 
    which cannot reproduce high-frequency appearance details like fine textures and sharp boundaries.

    Previous ray marching based networks \cite{lombardi2019neural,mildenhall2020nerf} 
    usually consider viewing directions as part of the input and directly output radiance at each point. 
    In contrast, we let the networks infer local shading normal and reflectance properties without knowing any viewing directions, 
    and then compute the radiance from the shading properties with given light-view directions according to a specified differentiable reflectance model.
    Essentially, we avoid inferring the challenging view- and light-dependent shading effects through deep neural processing as is done in previous work \cite{lombardi2019neural,mildenhall2020nerf}.
    Instead, we decouple the shading components and leverage classical reflectance models to regularize the neural rendering process;
    this achieves meaningful appearance modeling that is independent of light and view, 
    which yet can be naturally supplied for rendering under any lighting and viewpoint with complex specularities and other shading effects (see Fig.xxx).

}

\section{Related work}
\label{sec:related}

\paragraph{Neural scene representations.}
Previous work has applied deep neural networks to many 3D tasks with scene geometry modeled by various representations, such as volumes \cite{ji2017surfacenet,richter2018matryoshka}, 
point clouds \cite{qi2017pointnet}, implicit functions \cite{mescheder2018occupancy,sitzmann2019scene}, etc.
Reflectance modeling has also been explored with neural networks \cite{kuznetsov2019learning,vicini2019learned,rainer2019neural}.
We present the novel neural reflectance field that models both geometry and reflectance in a real scene.

Thies et al.~\shortcite{thies2019deferred} apply neural textures for realistic image synthesis,
but require a pre-acquired mesh as input.
Many previous works aim to do view synthesis without any known geometry.
Multiplane images
have been used in small-baseline view synthesis \cite{zhou2018stereo,srinivasan2019pushing}; 
however, such a view-dependent representation only supports limited viewing range and requires special fusion techniques to extend the range \cite{mildenhall2019local}.
Recent works leverage view-independent volumes, which are able to handle complex view-dependent effects \cite{sitzmann2019deepvoxels,lombardi2019neural}.
Our neural reflectance field models complete scene appearance; 
in addition to view synthesis, ours can also be used for other applications such as relighting. 

Recently, ray marching has been used to train many neural scene representations for view synthesis without any ground-truth 3D representations
\cite{mildenhall2020nerf,lombardi2019neural,sitzmann2019scene}.
Lombardi et al.~\shortcite{lombardi2019neural} apply ray marching in a discrete volume with a warping field for view synthesis. 
To make it generalizable to relighting, Bi et al.~\shortcite{bi2020deep} reconstruct discrete reflectance volumes with explicit per-voxel BRDFs;
however, the fixed resolution of the discrete volume limits the appearance details in the rendering.
In contrast, we leverage a \textit{continuous} functional neural representation and achieve much better results (see Fig.~\ref{fig:comp}).
Mildenhal et al.~\shortcite{mildenhall2020nerf} present a neural radiance field, which also represents a scene as a continuous function with a MLP. However, their representation only supports view synthesis by directly rendering radiance from a new viewpoint under fixed illumination.
We leverage a novel reflectance-aware ray marching framework
and learn to regress multiple decomposed shading components, which enables relighting and many other images synthesis applications.

\paragraph{Geometry and reflectance capture.}
Classically, modeling and re-rendering a real scene requires full reconstruction of its geometry and reflectance.
From captured images, scene geometry is usually reconstructed by structure-from-motion and multi-view stereo (MVS) 
\cite{kutulakos2000theory,esteban2004silhouette,furukawa2009accurate,schoenberger2016sfm,schoenberger2016mvs}, 
which have recently been extended using deep learning techniques \cite{yao2018mvsnet,yao2019recurrent,chen2019point,cheng2019deep}.

Reflectance acquisition traditionally requires sophisticated devices to sample the light-view space 
\cite{foo1997gonioreflectometer,matusik2003data,nielsen2015optimal,xu2016minimal,kang2018efficient,kang2019learning}.
Recently, many works use a practical device -- a modern cellphone that has a camera and a built-in flash light -- 
and capture flash images to acquire spatially varying BRDFs \cite{aittala2016neural,aittala2015two,hui2017reflectance,nam2018practical}.
While such a device only acquires reflectance samples under collocated light and view,
with enough samples, it is still sufficient to reconstruct many standard analytic reflectance models that are governed by the half-angle vector \cite{nam2018practical,hui2017reflectance}.
More recently, deep learning methods have made BRDF acquisition with a single flash image possible \cite{li2018materials,li2018learning,deschaintre2018single}.
Bi et al.~\shortcite{bi2020deep3d} extend the single-view case to a structured multi-view configuration, 
and reconstruct meshes with per-vertex BRDFs from only six images.

We aim to model geometry and appearance of complex real scenes from multi-view unstructured flash images.
From such inputs, Nam et al.~\shortcite{nam2018practical} leverage an initial mesh from MVS and reconstruct per-vertex BRDFs via traditional optimization.
However, it is very difficult for traditional mesh-based methods to recover challenging thin structures and sharp specularities of complex real scenes.
In this work, we address these issues by proposing a novel neural reflectance field to implicitly model the scene's geometry and reflectance,
bypassing explicit mesh reconstruction.
Our approach achieves photo-realistic rendering results with high-frequency appearance details that are significantly better than previous works.

\paragraph{Relighting and view synthesis.}
Scene acquisition and rendering can be also achieved using image-based techniques 
without explicit reconstruction \cite{levoy1996light,debevec2000acquiring}.
Recently, many learning based view synthesis methods have been presented \cite{zhou2018stereo,hedman2018deep,srinivasan2017learning,xu2019deep,mildenhall2020nerf}.
We extend the ray marching in the view synthesis works 
to a more general reflectance-aware ray marching framework,
which can also be used to do relighting.
Learning-based relighting methods have also been presented \cite{ren2015image,xu2018deep}, 
which are able to reproduce challenging appearance effects.
Many techniques regress images under novel lighting from sparse inputs without any explicit geometry reasoning \cite{xu2018deep,zhou2019deep,sun2019single}, 
but are unable to recover accurate hard shadows.
Philip et al.~\shortcite{philip2019multi} require a mesh from MVS for shadowing computation.
Our network learns to regress volume density to model detailed scene geometry.
Our ray marching considers light transmittance in ray integration, which recovers challenging hard shadows.

Note that, previous image-based techniques often send viewing \cite{mildenhall2020nerf,lombardi2019neural} 
or lighting \cite{xu2018deep,sun2019single} information 
as additional inputs to the network, and compute challenging view- or light- dependent shading effects through the network processing.
In contrast, we leverage classical reflectance models to regularize the learning process;
our neural reflectance field is independent of the viewing and lighting directions,
and we use the regressed reflectance and normal to compute shading under any lighting and viewpoint.
Our approach can properly model scene appearance and reproduce challenging view-dependent and light-dependent shading effects.

\section{Reflectance-aware Ray Marching}
\label{sec:raymarchingframework}
While differentiable ray marching has been used in recent works \cite{lombardi2019neural,mildenhall2020nerf},
these methods focus on view synthesis and only consider view-dependent effects.
We utilize classical reflectance models in a more general ray marching framework (see Fig.~\ref{fig:overview}) that also models lighting and enables relighting and other re-rendering applications.
Our reflectance-aware rendering framework is differentiable 
and can be easily combined with deep learning to learn scene appearance.
In this section, we first discuss the underlying rendering equation that governs our volume rendering (Sec.~\ref{sec:volumerendering}), and then introduce our ray marching framework that numerically computes the equation in a differentiable way (Sec.~\ref{sec:raymarching}). 

\subsection{Rendering equation}
\label{sec:volumerendering}
In general, for non-emissive and non-absorptive volumes, physically-based volume rendering is governed by the volume rendering equation \cite{novak2018monte} that computes the radiance $\Radiance(\EPos, \EDir)$ at point $\EPos$ in direction $\EDir$:
\begin{align}
    \Radiance(\EPos, \EDir) & = \int_0^\infty \Trans_{\EPos}(\PosX) \Extinct(\PosX) \Radiance_s(\PosX,\EDir) d\posx,\label{eqn:render}\\
   \text{where\ } \Trans_{\EPos}(\PosX) &= e^{-\int_0^\posx \Extinct(\EPos-u\EDir) du} .\label{eqn:trans}
\end{align}
Here, $\posx$ represents the 1D location on a ray traced in the volume, 
$\PosX = \EPos-\posx\EDir$ represents the 3D point at $\posx$, the point $\EPos$ typically represents the camera location, and $\Radiance_s(\PosX,\EDir)$ 
represents the scattered light at $\PosX$ along $\EDir$. 
$\Extinct$ is the extinction coefficient that indicates the probability density of medium particles;
we refer to $\Extinct$ as volume density in this paper.
$\Trans_{\EPos}(\PosX)$ represents the transmittance factor which determines the loss of light along the ray from $\EPos$ to $\PosX$.

Eqn.~\ref{eqn:render} computes the radiance that arrives 
at $\EPos$ by integrating the modulated in-scattered light $\Radiance_s$ along the ray,
\begin{equation}
    \Radiance_s(\PosX,\EDir) =  \int_{\mathcal{S}} f_p(\PosX,\EDir,\LDir) \Radiance_i(\PosX,\LDir) d\LDir,
    \label{eqn:phaseInt}
\end{equation}
where $\mathcal{S}$ is a unit sphere, $f_p$ is a phase function that governs light scattering, 
and $\Radiance_i(\PosX,\LDir)$ is the incident radiance arriving at $\PosX$ from direction $\LDir$.

Note that previous work \cite{mildenhall2020nerf} directly encodes $\Radiance_s$ without considering any form of Eqn.~\ref{eqn:phaseInt}; 
this assumes fixed illumination and only works for view synthesis.
In contrast, we consider single-bounce direct illumination under a single point light source to approximate $\Radiance_s$.
Inspired by \cite{max1995optical},
we compute $\Radiance_s$ with an explicit reflectance term that assumes the role of a phase function:
\begin{equation}
    \Radiance_s(\PosX,\EDir) =  f_r(\PosX,\EDir,\LDir, \SNormal(\PosX), \SReflect(\PosX)) \Radiance_i(\PosX,\LDir),
    \label{eqn:shading}
\end{equation}
where $f_r$ represents a differentiable reflectance model with parameters $\SReflect$,
$\SNormal$ is the local surface shading normal, and $\Radiance_i$ represents the incident radiance as in Eqn.~\ref{eqn:phaseInt}.
When only considering direct illumination from a point light source, 
$\Radiance_i$ is determined by the intensity of the light source and the loss of light due to extinction through the volume:
\begin{equation}
    \Radiance_i(\PosX,\LDir) =  \Trans_{\LPos}(\PosX) \Radiance_{\LPos}(\PosX),
    \label{eqn:incident}
\end{equation}
where
$\Trans_{\LPos}$ is the transmittance from the light to the shading point, 
and $\Radiance_{\LPos}$ represents the light intensity with the consideration of distance attenuation.
Here, $\LPos$ denotes the position of the point light source, and thus $\LDir$ corresponds to the direction of the vector $\LPos-\PosX$.

Therefore, by combining Eqn.~\ref{eqn:render}, \ref{eqn:shading}, \ref{eqn:incident}, 
our volume rendering is governed by a \emph{reflectance-aware} rendering equation \cite{max1995optical,kniss2003model}: 
\begin{equation}
    \Radiance(\EPos, \EDir) = \int \Trans_{\EPos}(\PosX)\Trans_{\LPos}(\PosX) \Extinct(\posx)  f_r(\PosX,\EDir,\LDir, \SNormal(\PosX), \SReflect(\PosX)) \Radiance_{\LPos}(\PosX)d\posx.
    \label{eqn:refrendering}
\end{equation}
Our equation considers complete one-bounce camera-volume-light paths in the light transport.
Unlike previous work that only considers the view transmittance or opacity between the shading point and the camera \cite{mildenhall2020nerf,lombardi2019neural},
we also explicitly express the light transmittance ($\Trans_{\LPos}$) from the point to the light,
which allows us to render realistic shadows under different point light sources.
Essentially, instead of modeling scene \emph{radiance} $\Radiance_s$ as is in previous view synthesis work \cite{mildenhall2020nerf},
we decouple the multiple factors ($\Trans_{\LPos}$, $f_r$, $\Radiance_{\LPos}$) that are embedded in $\Radiance_s$, 
and explicitly model the scene \emph{reflectance} parameters in $f_r$, 
thus allowing for reflectance-aware rendering for both view synthesis and relighting with realistic shading and shadowing effects.

\subsection{Ray marching}
\label{sec:raymarching}
We use ray marching to numerically estimate the continuous integral in Eqn.~\ref{eqn:refrendering}
similarly to prior work on volume rendering \cite{max1995optical,kniss2003model}; this is illustrated in Fig.~\ref{fig:overview}.
Specifically, we march rays from the camera center through each pixel on the image plane and 
sample a sequence of $\EN$ shading points $\PosX_j$ on each ray.
The rendering equation can be estimated by:
\begin{equation}
    \Radiance(\EPos, \EDir) = \sum_{j=0}^{\EN} \Trans_{\EPos}(\PosX_j)\Trans_{\LPos}(\PosX_j) (1-\exp(-\Extinct(\PosX_j)\Delta \posx_j)) f_r(\PosX_j) \Radiance_{\LPos}(\PosX_j),
    \label{eqn:numrendering}
\end{equation}
where $\Delta \posx_j$ represents the ray step size at point $\PosX_j$. Here, we omit the other parameters ($\EDir$, $\LDir$, $\SNormal$, $\SReflect$) in $f_r$ for brevity.
Here, $\Trans_{\EPos}(\PosX_j)$ is also an integral (Eqn.~\ref{eqn:trans}) and can be numerically evaluated by:
\begin{equation}
    \Trans_{\EPos}(\PosX_j) = \exp \left( -\sum_{k=0}^j \Extinct(\PosX_k)\Delta \posx_k \right).
    \label{eqn:numETrans}
\end{equation}
The transmittance $\Trans_{\LPos}(\PosX_j)$ can be similarly evaluated,
but it requires sampling another sequence of points $\PosXPr_p$ on an additional ray marched from the light source to the shading point $\PosX_j$:
\begin{equation}
    \Trans_{\LPos}(\PosX_j) = \exp \left( -\sum_p \Extinct(\PosXPr_p)\Delta \posxPr_p \right).
    \label{eqn:numLTrans}
\end{equation}
Naively computing Eqn.~\ref{eqn:numLTrans} for Eqn.~\ref{eqn:numrendering} would require marching a large number of light rays for all shading points on all camera rays.
Instead, we leverage a collocated light source and camera setup (where the camera and light rays are the same) to avoid this during training; this is described in Sec.~\ref{sec:learningreflectance}. 
At inference time we precompute an adaptive transmittance volume to efficiently approximate Eqn.~\ref{eqn:numLTrans} under any point light; this is described in Sec.~\ref{sec:renderingreflectance}.

Equations \ref{eqn:numrendering}, \ref{eqn:numETrans}, \ref{eqn:numLTrans} express our reflectance-aware ray marching framework.
Given a camera $\EPos$ and a point light $\LPos$,
the framework computes the radiance of any marching ray through a scene from 
the volume density $\Extinct$, normal $\SNormal$, 
and reflectance properties $\SReflect$ of the points in the scene.
\cite{bi2020deep} utilizes a rendering equation similar to ours (Eqn.~\ref{eqn:refrendering}), 
but they leverage classical opacity accumulation -- where the opacity is given by $\alpha = 1-\exp(-\Extinct\Delta\posx)$ -- to numerically evaluate the integral,
which only supports a fixed step size for ray marching.
In contrast, we utilize volume density $\Extinct$ for numerical estimation,
which is more general and allows the step sizes to vary across the shading points.
This enables applying better adaptive sampling strategies to distribute the shading points along both camera and light rays  (See \ref{sec:learningreflectance} and Sec.~\ref{sec:renderingreflectance}).
In addition, since volume density is standard in Monte Carlo based volume rendering,
the model learned from our ray marching framework can be also used in standard rendering engines 
for general graphics applications (See. Fig.~\ref{fig:mituba}). 

Our ray marching framework supports any differentiable reflectance model $f_r$,
which makes the full rendering process trivially differentiable.
In this work, we demonstrate most results using a classical analytic BRDF \cite{karis2013real} for $f_r$,
which models the reflectance of opaque surfaces with a diffuse albedo and a specular roughness.
We also show results with hair/fur reflectance models \cite{kajiya1989rendering} that model the appearance of furry objects,
demonstrating the generality of this formulation.
Our reflectance-aware ray marching framework can potentially be combined with any module 
that is able to provide the rendering properties ($\Extinct$, $\SNormal$, $\SReflect$) of an arbitrary point in the scene.
In this work, we use a neural network to regress the necessary rendering properties.

\section{Neural reflectance fields}
We now present our \nrf{} representation that uses deep fully connected networks to model scene geometry and reflectance (Sec.~\ref{sec:network}).
As shown in Fig.~\ref{fig:overview}, this network can be used in conjunction with the reflectance-aware ray marching scheme described previously.
We show how it can effectively trained from cellphone flash images (Sec.~\ref{sec:learningreflectance}).
We also present an adaptive transmittance volume for light transmittance precomputation,
enabling efficient rendering under any novel light and view positions with realistic shadows (Sec.~\ref{sec:renderingreflectance}).

\subsection{Network}
\label{sec:network}
Given a reflectance model $f_r$ with $m$ parameters,
a \nrf{} outputs a $(4+m)$-dimensional vector---comprising volume density $\Extinct$ (1-D), normal $\SNormal$ (3-D) and reflectance properties $\SReflect$ ($m$-D)---at any 3-D position $\PosX = (x,y,z)$ in a scene. In practice, we use a microfacet BRDF model~\cite{Walter2007} where $\SReflect$ comprises diffuse albedo and specular roughness, though we also demonstrate an extension using a fur reflectance model~\cite{kajiya1989rendering}.
We parameterize \nrfs{} using an MLP with 14 fully connected layers and ReLU activation layers. Please refer to the supplementary material for the detailed architecture of our network.
 
Inspired by \cite{rahaman2018spectral,vaswani2017attention,mildenhall2020nerf}, 
we use a frequency-based positional encoding of a given 3D location $\PosX = (x,y,z)$.  
In particular, given each dimension of the 3D $(x,y,z)$ point, we map the scalar value $v$ to
\begin{equation}
    \gamma(v) = (\sin (2^0\pi v), \cos(2^0\pi v), \ldots, \sin (2^{W-1}\pi v), \cos(2^{W-1}\pi v)),
\end{equation} 
where $W$ represents the highest frequency level ($W=10$ in our experiments). These are input to the MLP to regress the scene properties at the encoding $\gamma(x)$, $\gamma(y)$ and $\gamma(z)$.

Unlike~\cite{mildenhall2020nerf}, which also use a positional encoding of the viewing direction,
we only use the 3D location as input, inferring view- (and light-) independent scene appearance properties.
This is possible because we separate out these factors and compute shading, viewing, and lighting information in our ray marching framework (Eqn.~\ref{eqn:refrendering},~\ref{eqn:numrendering}); this allows \nrfs{} to be directly plugged into it for high-quality rendering.

\subsection{Learning \nrfs{} from flash images}
\label{sec:learningreflectance}

We now describe how we can use \nrfs{} to reconstruct the appearance of a real-world scene from images.
Each \nrf{} is fit to a specific scene via a training process.
Since our whole rendering process (the representation and the ray marching) is differentiable, we train the \nrf{} network to minimize the  error between rendered images and captured images of the scene.

\paragraph{Collocated light and view.} In particular, we capture flash images with collocated light and view to train our networks. 
Such images can be easily captured by a cellphone with a camera and a flash.
The collocated setting leads to $\EPos = \LPos$ in our training.
One key benefit of using collocated light and view is that the view transmittance $\Trans_{\EPos}$ and the light transmittance $\Trans_{\LPos}$ become equal 
in Eqn.~\ref{eqn:numrendering}.
This avoids marching an additional ray towards the light at every shading point, 
which would make training intractable.
This capture setup thus has the advantage of making both \emph{acquisition} and \emph{training} practical.
However, this also means that our input images represent an extremely sparse sampling of scene appearance across the view-light space. 
In fact, we have no samples of the scene for lighting from any non-zero angle with respect to the camera.
In spite of this, we show that we can reconstruct high-quality scene appearance and render images under arbitrary view and even non-collocated lighting.


\paragraph{Adaptive sampling for camera rays.} To optimize the point sampling in ray marching,
for each scene, we train two networks---a coarse and a fine neural reflectance field---and render using a coarse-to-fine adaptive sampling procedure.
Inspired by \cite{mildenhall2020nerf}, we first sample a sparse set of points on each marching ray with stratified sampling 
to compute a distribution function using the coarse network,
then sample a dense set of points from the distribution function to compute the final radiance value using the fine network.

In particular, we divide each full ray segment into $\EN_1$ bins and randomly sample a point from each bin to get stratified samples.
From these points, we can compute the radiance from the coarse-level network for the ray using Eqn.~\ref{eqn:numrendering}.
As a side product, we can also produce corresponding per-point contribution weights
\begin{equation}
    \Import(\PosX_j) = \Trans_{\EPos}(\PosX_j) (1-\exp(-\Extinct(\PosX_j)\Delta \posx_j)).
    \label{eqn:importEye}
\end{equation}
The weight $\Import(\PosX)$ essentially describes how visible the point at $\PosX$ is to the camera.
We construct a piece-wise constant probability distribution by normalizing the per-point weights $\Import(\PosX_j)$ and then sample $\EN_2$ points from this distribution,
which adaptively selects new samples according to the visibility information gathered from the coarse neural reflectance field.
We then use all $\EN=\EN_1+\EN_2$ sampled points to compute the final radiance with Eqn.~\ref{eqn:numrendering} using the rendering parameters from the fine reflectance field.
This coarse-to-fine adaptive sampling effectively distributes more sampled points in the regions that contribute most to the rendering integral,
allowing for accurate shading computation with high-frequency details.

\subsection{Efficient rendering under novel light and view}
\label{sec:renderingreflectance}
While our neural reflectance field is learned from flash images with collocated light and view,
the learned representation can be directly used to render the scene with single-scattering effects using any light and view positions with Eqn.~\ref{eqn:numrendering}.
However, accurately computing $\Trans_{\LPos}$ at inference time under
novel non-collocated light and view (unlike training) is extremely computationally expensive.
Therefore, we propose to pre-compute an adaptive transmittance volume to effectively approximate $\Trans_{\LPos}$.
\begin{figure}[t]
    \includegraphics[width=\linewidth]{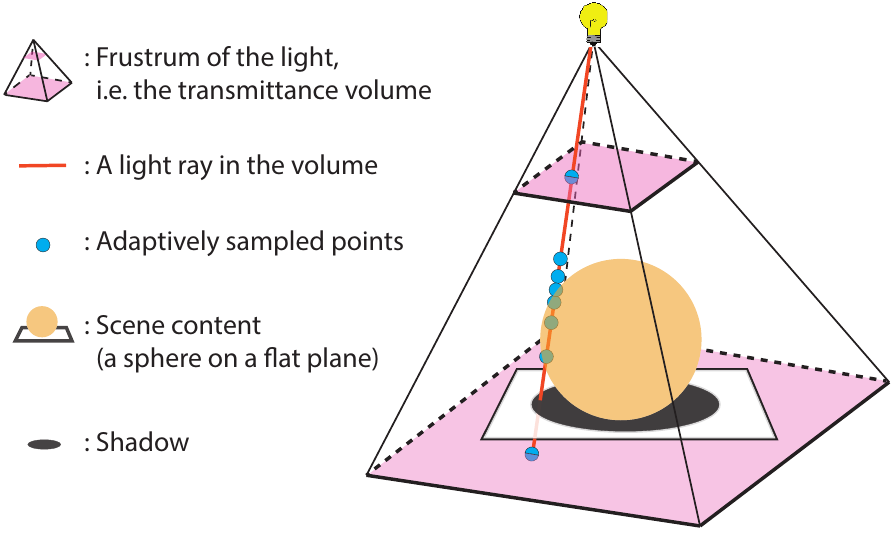}
    \caption{Adaptive transmittance volume. We pre-compute a light transmittance volume within the frustrum of a virtual view at the point light source. We leverage a coarse-to-fine strategy to adaptively distribute the sampled points (blue) around the visible scene structures along each light ray (red),
    enabling an efficient transmittance representation.}
    \label{fig:atv}
\end{figure}

\paragraph{Adaptive sampling for light transmittance volume.}
Inspired by the classical shadow map technique \cite{stamminger2002perspective,williams1978casting} in rasterization,
we use our learned neural reflectance field to compute a transmittance volume similar to \cite{lokovic2000deep} for fast light transmittance computation.
Specifically, we place a virtual image plane in front of the point light source towards the scene
and march a ray through each pixel, 
analogously to what is classically done for ray marching from the camera.
Similar to the \textit{adaptive sampling for camera rays} described in Sec.~\ref{sec:learningreflectance},
we use the two trained networks (a coarse and a fine network) to perform adaptive sampling. 
We first utilize the coarse representation to compute a visibility-aware distribution function using sparse points sampled from stratified bins; 
we then sample dense points from the distribution.
We combine the samples from both passes and compute their light transmittance,
resulting in a transmittance volume that adapts to the visibility information inferred from the coarse network.
This adaptive transmittance volume is illustrated in Fig.~\ref{fig:atv}. 

\paragraph{Final rendering.}
We do ray marching from the viewpoint to render an image under any point light source from any viewpoint using the learned network and the pre-computed adaptive transmittance volume.
At any given shading point, we locate the nearest sampled points and then linearly interpolate the transmittance volume to get the required light transmittance, similar to \cite{lokovic2000deep}.
This allows for realistic shadowing effects to be well recovered in our results when doing relighting.

We also apply coarse-to-fine sampling on the camera rays for the rendering at inference, 
as we described in Sec.~\ref{sec:learningreflectance} at training.
Basically, at inference, we apply coarse-to-fine adaptive sampling in ray marching from both the light and the camera,
which achieves efficient light transmittance computation and effective final image synthesis.

\begin{figure*}[t]
    \includegraphics[width=\linewidth]{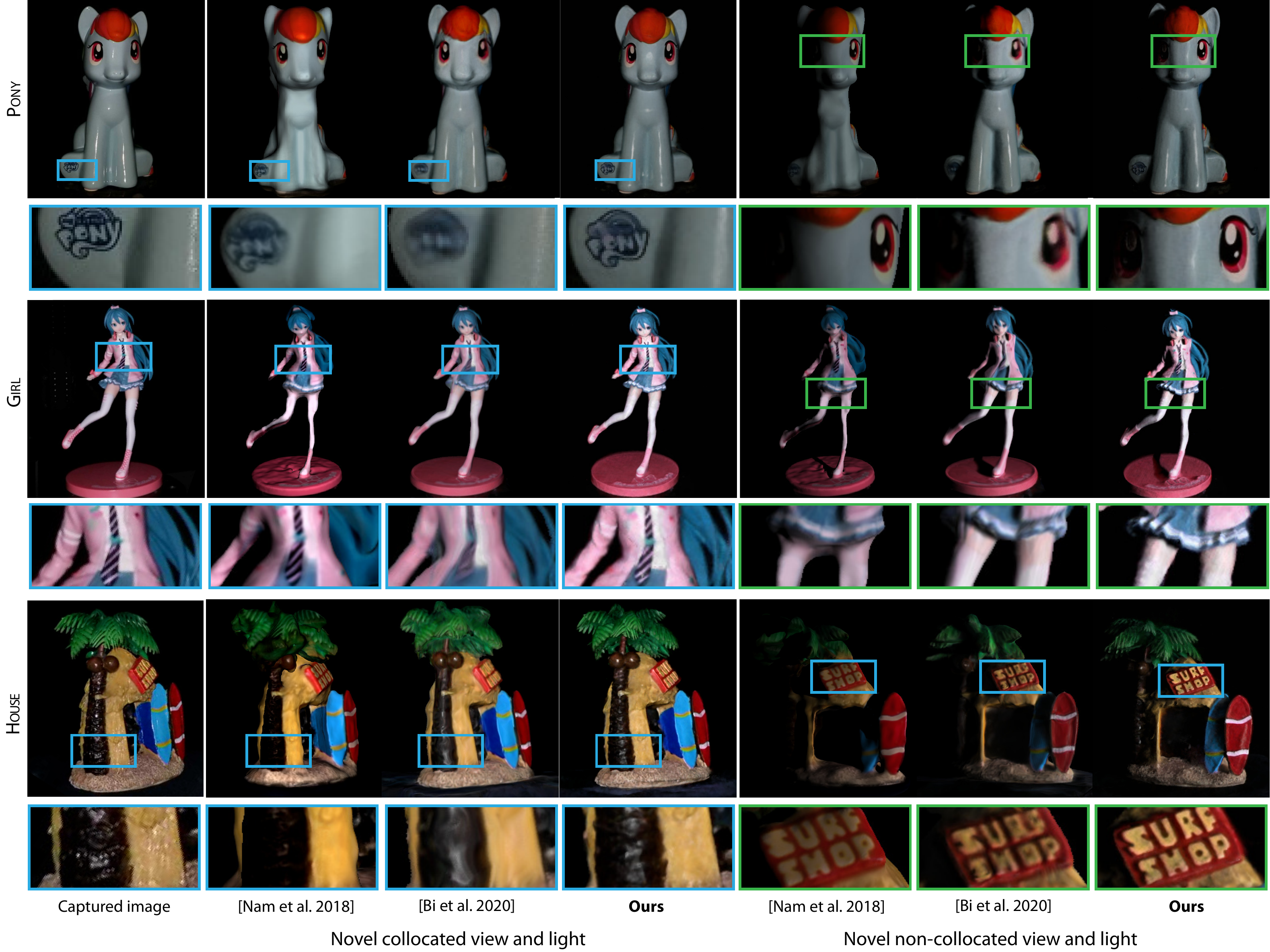}
    \caption{Comparisons with previous work. We compare our view synthesis and relighting results with a state-of-the-art mesh-based method \cite{nam2018practical} and a previous learning-based method \cite{bi2020deep} on complex real scenes. We show one captured image (not used for training) on the left. We compare re-renderings under novel collocated camera and light (middle) and novel \emph{non-collocated} camera and light (right). As can be seen here, even a state-of-the-art mesh reconstruction method fails to accurately reconstruct complex real-world scenes. While the learning-based approach improves on this result, it produces blurry results. In contrast, our method produces realistic results with high-frequency textures, specular highlights, and complex shadowing.} 
    \label{fig:comp}
\end{figure*}

As noted before, during training, our network only sees images that are captured under collocated light and view and do not have any shadows.
Yet, our method is able to learn a volume density that meaningfully expresses scene geometry.
This allows us to synthesize high-quality relighting and view synthesis results with realistic shadows, specularities and other appearance effects under novel, non-collocated light and view, as illustrated in Figs.~\ref{fig:teaser}, \ref{fig:comp}, \ref{fig:more_results}.

\section{implementation}
\label{sec:implementation.}

\paragraph{Data acquisition.}
As discussed in Sec.~\ref{sec:learningreflectance}, we reconstruct \nrfs{} from images captured under collocated view and lighting.
Such data can be practically acquired by shooting a video using a handheld cellphone with flash. 
We show acquisition and rendering results of one human portrait using this handheld setup in Fig.~\ref{fig:face}; in this case, we selected 150 frames from the video as input.
To facilitate the data acquisition, for other results, we use a robotic arm holding a cellphone to automatically capture scenes that are composed of different static objects.
We capture about 400 images using this automatic setup.
We use a Samsung Galaxy Note 8 to capture all our real scenes.
The camera parameters are calibrated using structure from motion in COLMAP \cite{schoenberger2016sfm}.
Our method does not require accurate background masks for the input images to train the network.
We simply crop the central regions around the objects in the captured images to avoid training on too many background pixels.
Each network is trained in a scene-dependent way, using the input images for that single scene.

\paragraph{Reflectance model.}
Our representation works with any differentiable reflectance model. In practice, we use a microfacet BRDF model that combines a diffuse Lambertian term with a specular term that uses the GGX distribution \cite{Walter2007}. The parameters of this model include a diffuse albedo and a specular roughness. With this model, the \nrf{} MLP thus outputs a 8-D vector at every scene point, corresponding to the 3-D diffuse albedo, 1-D specular roughness, 3-D surface normal and 1-D transmittance. We use this BRDF model for every result in the paper, except for Fig.~\ref{fig:fur} where we capture a furry object. Here we use the classical fur reflectance model \cite{kajiya1989rendering} and replace the surface normal $\bm{n}$ with a fiber tangent vector. 

\paragraph{Training parameters and loss function.}
We implement our neural reflectance field and ray marching in PyTorch.
During training, we randomly sample $50\times 50$ pixel rays as a batch to train our network under collocated light as described in Sec.~\ref{sec:learningreflectance}.
We use Adam optimizer with an initial learning rate of 0.0001.
We use $\EN_1=64$ coarse samples and $\EN_2=128$ fine samples to adaptively sample light rays when building the adaptive transmittance volume and camera rays when computing the final radiance.

We supervise the regressed radiance values from both the coarse and the fine network 
with the ground truth radiance $\tilde{\Radiance}$ from the captured images using the $L_2$ loss.
Since we consider opaque objects, 
we also regularize the ray transmittance (from the fine network), forcing it to be close to 0 or 1, which is helpful to get a clean background.
Our total loss function is given by:
\begin{equation}
    \sum_q \| \Radiance^q_{\text{coarse}} - \tilde{\Radiance}^q\|^2 + \| \Radiance^q_{\text{fine}} - \tilde{\Radiance}^q\|^2 + \beta[\log(\Trans^q_{\EPos})+\log(1-\Trans^q_{\EPos})],
\end{equation}
where $q$ denotes a pixel ray and $\beta = 0.0001$ is a hyper-parameter that controls the strength of the regularization term.

\paragraph{Run time.}
We use 4 NVIDIA RTX 2080Ti GPUs to train each reflectance field network for about 2 days.
At inference time, the network takes about 30 seconds to render a $512\times 512$ image using our adaptive transmittance volume.

\section{Results}
\label{sec:experiments}
We now demonstrate our results in this section.
We first evaluate our method by comparing our view synthesis and relighting results with other methods. 
We then show more results and applications of our method.
Please refer to the supplementary video for more video results.

\begin{figure*}[t]
    \includegraphics[width=\linewidth]{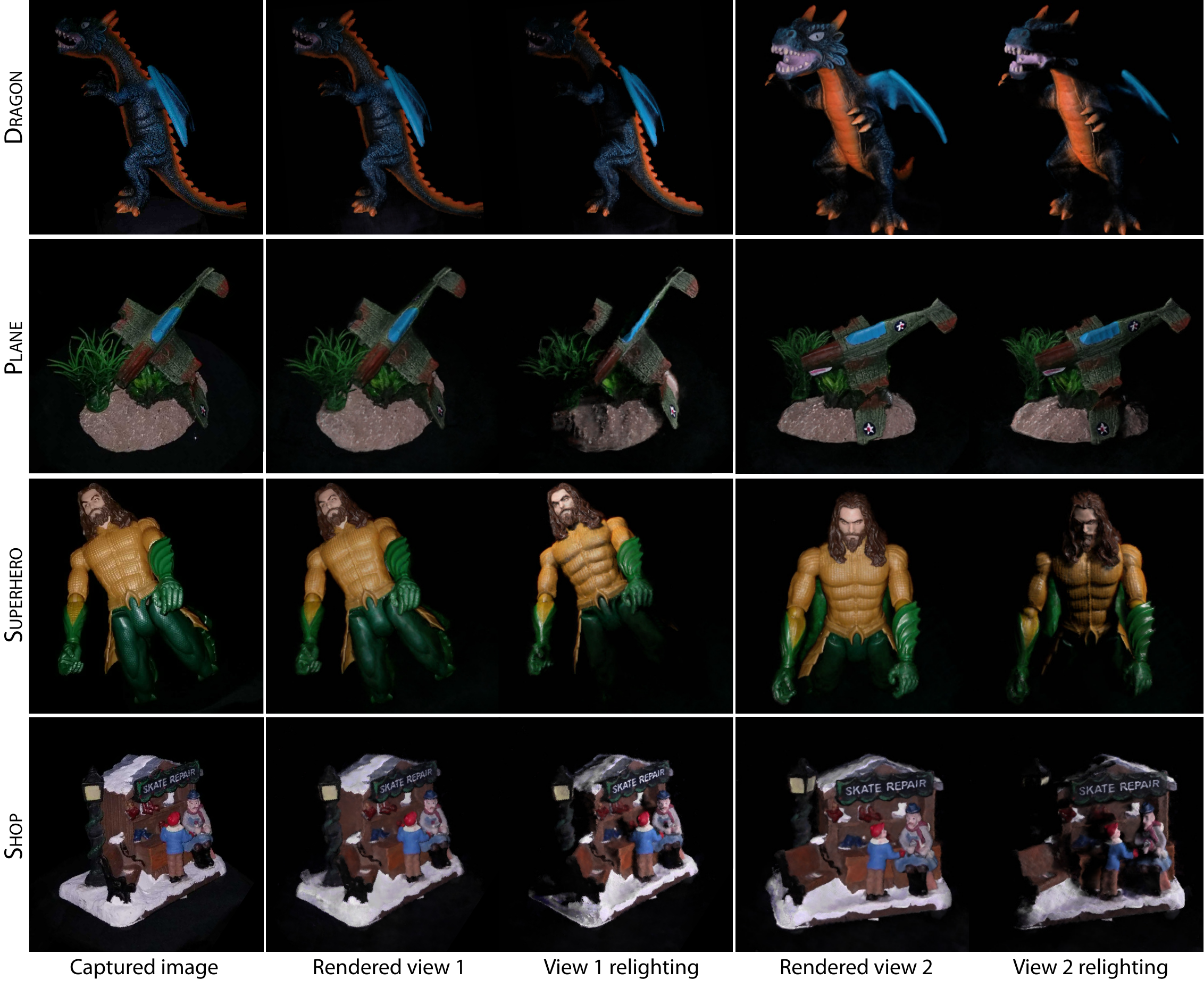}
    \caption{Additional view synthesis and relighting results of our method. We show our rendered images under collocated light-view settings from two different views (columns 2 and 4). We also show captured image (not used for training) from view 1 in the left most column to demonstrate that our renderings closely reproduce the ground truth appearance of these scenes. We also demonstrate relighting results (columns 3 and 5) from each view, in which the light and view are no longer collocated, leading to challenging cast shadows.}
    \label{fig:more_results}
\end{figure*}

\begin{figure}[t]
    \includegraphics[width=\linewidth]{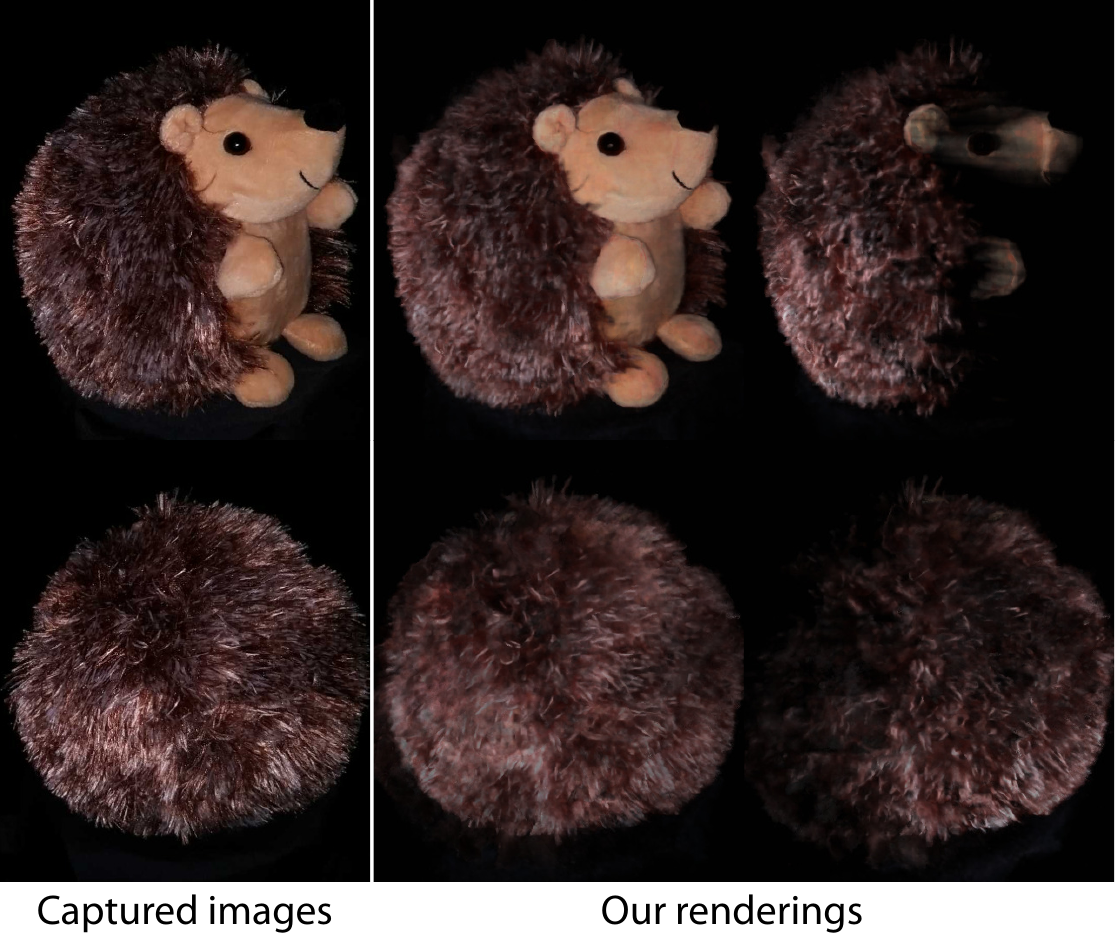}
    \caption{Results on a furry object. We incorporate a fur reflectance model \cite{kajiya1989rendering} in our representation to capture this object. We show examples of ground truth captured images (that were not used for training) on the left, and renderings from our method on the right.}
    \label{fig:fur}
\end{figure}

\begin{figure}[t]
    \includegraphics[width=\linewidth]{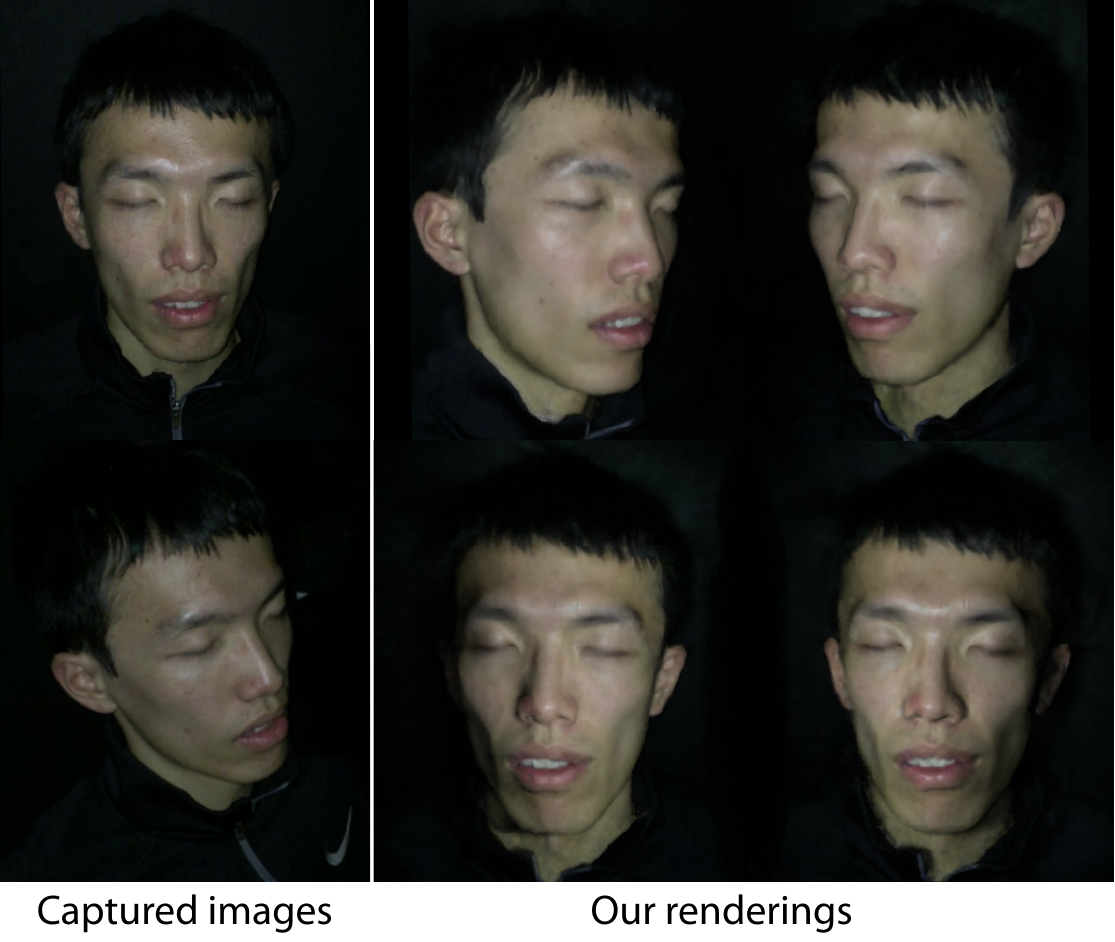}
    \caption{Results on a human face. We show examples of ground truth captured images (that were not used for training) on the left, and on the right show renderings from our method for novel view synthesis with a collocated light (top) and relighting with a non-collocated light (bottom).}
    \label{fig:face}
\end{figure}

\begin{figure}[t]
    \includegraphics[width=\linewidth]{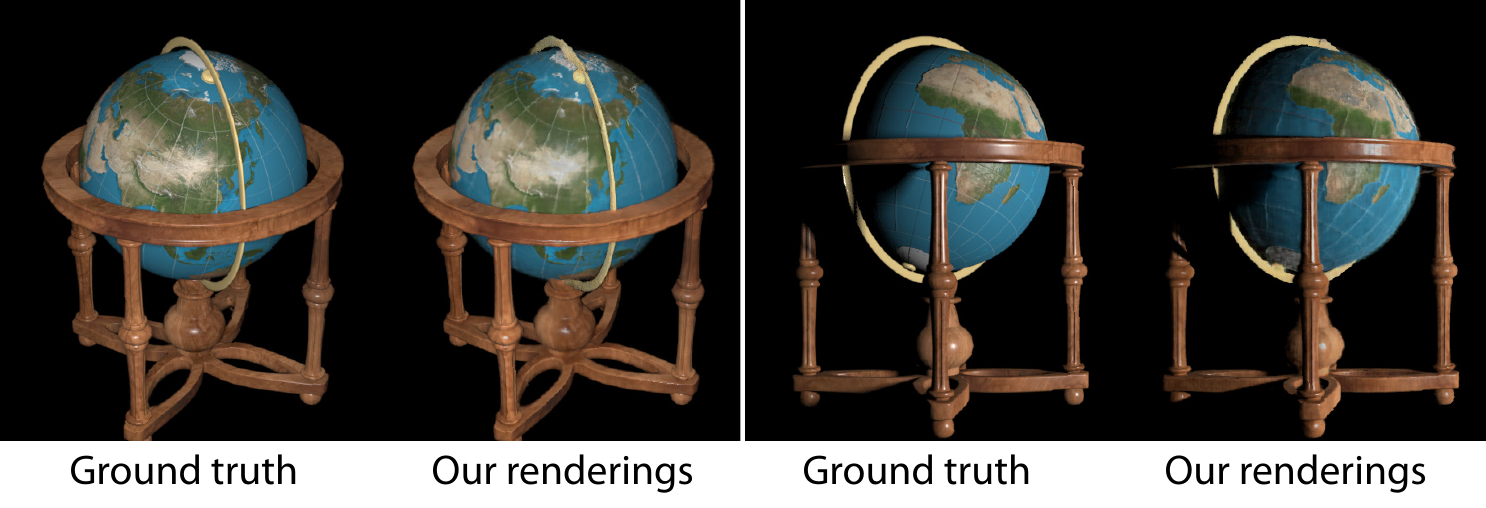}
    \caption{Synthetic results. We compare the rendered images of our method under novel lighting and viewpoint with the ground truth images.}
    \label{fig:syn}
\end{figure}

\begin{figure}[t]
    \includegraphics[width=\linewidth]{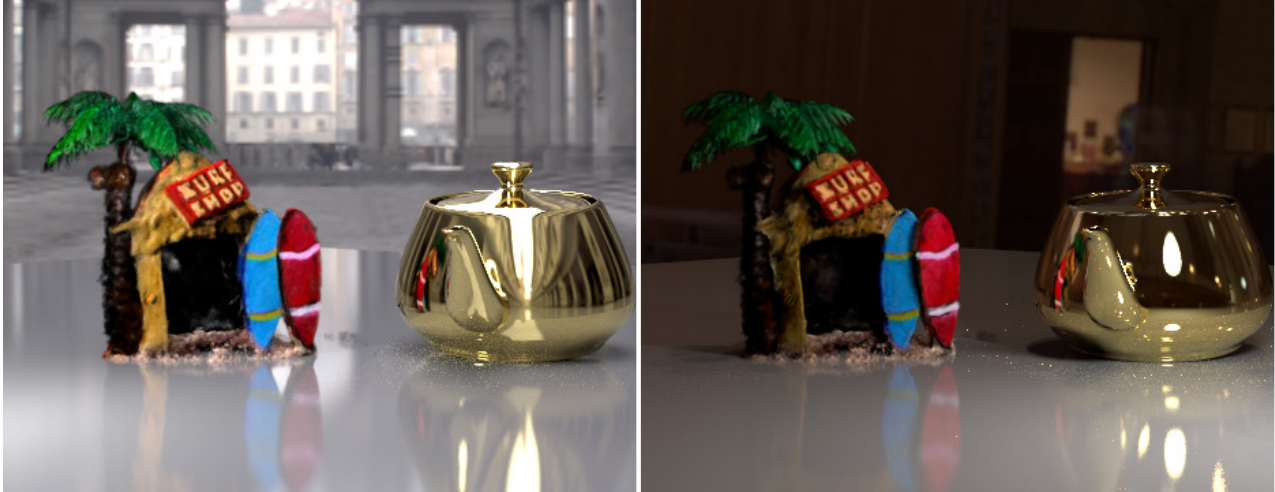}
    \caption{We show examples of combining the neural reflectance field of a real scene (\SHouse) with a traditional synthetic scene (teapot). We render the composed scene using a standard rendering engine (Mitsuba \cite{mitsuba}) under complex illumination.}
    \label{fig:mituba}
\end{figure}

\paragraph{Comparisons with previous methods.}
Most previous learning-based works focus only on the sub-problems of relighting \cite{xu2018deep,ren2015image} or view synthesis \cite{xu2019deep,mildenhall2020nerf,lombardi2019neural,mildenhall2019local}, and capture images with a fixed camera or fixed illumination, respectively. 
Instead, our input light and view are collocated and vary across all input images, allowing us to build a holistic scene representation that allows for both view synthesis and relighting.
We are aware of only a few methods that address this problem and we compare against two of them.
The first is a state-of-the-art mesh-based appearance acquisition method \cite{nam2018practical} that reconstructs a 3D mesh and per-vertex BRDFs from collocated flash images. The reconstructed geometry and reflectance can then be used to achieve relighting and view synthesis.
We also compare with a learning-based method \cite{bi2020deep}, 
that predicts a discrete volume with explicit per-voxel reflectance properties.
This technique supports relighting and view synthesis via opacity accumulation-based ray marching.
In Fig.~\ref{fig:comp}, we show qualitative comparisons of images renderer from the respective reconstructions under novel collocated and non-collocated light-view settings. 
Results for all methods were generated from the same inputs by their respective authors.
Please refer to the supplementary video for video comparisons.

Fig.~\ref{fig:comp} shows that our method achieves significantly better rendering results than \cite{nam2018practical}.
They leverage a classical multi-view stereo (MVS) method to reconstruct an initial mesh,
 and then recover a refined mesh and per-vertex BRDFs via traditional optimization.
However, for challenging real scenes, MVS often fails to recover reasonable initial geometry 
in regions that with little texture, high specularity, or thin structures.
This leads to highly distorted and even missing geometry in their results.
In addition, since specular effects typically influence very few pixels, their optimization-based reflectance estimation step is unable to recover them, leading to a mostly diffuse appearance,
.
In contrast, our neural reflectance field 
bypasses mesh reconstruction and is able to accurately resolve fine geometric structure with volume densities. This leads to high-quality rendering results with realistic geometric details, high specularities and hard shadows.

Our method also outperforms the previous deep volume rendering method \cite{bi2020deep}.
While that method also avoids the geometric reconstruction issues arising from Nam et al.~\shortcite{nam2018practical},
it fails to recover high-frequency details in the results, as reflected in many of the insets shown in Fig.~\ref{fig:comp}.
This is because they regress a discrete volume with per-voxel BRDFs;
the rendering quality is limited by the resolution of the volume, which is strictly constrained by system memory.
Instead, by leveraging a continuous functional representation, our network can properly recover high-frequency appearance.
Our neural reflectance field is also extremely compact, with weights consuming only ~5 MB of memory.
In contrast, \cite{bi2020deep} uses a network that requires ~400 MB to predict a volume that consumes several gigabytes of memory during rendering.
Our approach is more efficient in terms of memory usage and has more potential to be extended to capture of large-scale real scenes.

\paragraph{Additional results on diverse real scenes.}
We now demonstrate additional view synthesis and relighting results from our method on diverse real scenes in Figs.~\ref{fig:more_results}, \ref{fig:fur}, and \ref{fig:face}.
Fig.~\ref{fig:more_results} shows results on complex objects.
Our method successfully recovers various challenging high-frequency appearance effects, 
such as detailed geometry, complex textures, specularities, and hard shadows.
Note that the detailed thin geometry of the grass in \SPlane\ and the complex normal variation on the surfaces in \SDragon\ and \SAMan\ are all well reproduced realistically.
Our method can also handle challenging scenes that consist of multiple objects, like \SShop.
These lead to complex cast shadows between objects, that our method accurately reproduces in spite of never having observed them in the input images. This can be attributed to the ability of our method to infer reliable geometry (in the form of a volume density) from just collocated image samples. 

In Fig.~\ref{fig:fur}, we acquire the appearance of a furry object.
Here, we plug in the classical fur reflectance model \cite{kajiya1989rendering} into our representation, demonstrating the ability of \nrfs{} to work with a wide range of reflectance models.
While the results here are slightly blurrier than the other scenes (Fig.~\ref{fig:more_results}), they still look very realistic with the desired furry appearance. 
Our method can also be used to capture facial appearance, as shown in Fig.~\ref{fig:face}.
Here, we use a handheld cellphone and simply capture a video (with flash) walking around the person. 
From this video, we sample 150 images and train a \nrf{} that allows for re-rendering under varying viewpoint and lighting.
Acquiring facial appearance is an extensively studied problem and recent deep learning-based approaches have demonstrated portrait relighting from sparse inputs. 
However, these either require calibrated illumination \cite{xu2018deep,meka2019deep} or focus on low-frequency illumination \cite{sun2019single,zhou2019deep}.
In contrast, our images are captured with a practical setup, and are of high quality with realistic specularities and hard shadows,
in spite of not making any face-specific assumptions in our method. 

\paragraph{Synthetic results.}
Since our setup only captures images under collocated view and light, we do not have ground truth captured images to evaluate renderings under non-collocated camera and light.
We thus compare using a synthetic scene in Fig.~\ref{fig:syn}, where we can render the ground truth under any lighting and viewpoint.
As shown in Fig.~\ref{fig:syn}, our method is able to accurately reproduce the high-frequency textures, specularities, and hard shadows in the rendered images, which are very close to the ground truth.


\paragraph{Integrating with Monte-Carlo renderers.}
While neural rendering approaches have made remarkable progress in the recent past, one challenge with them is that they still require custom components that may not be consistent with standard scene representations and rendering engines.
In addition, most current methods focus on the view synthesis task \cite{mildenhall2020nerf,lombardi2019neural} and do not model the interaction of lighting with the captured scene.
While Bi et al.~\shortcite{bi2020deep} do model lighting, it is based on opacity accumulation and only supports a fixed step size, which is not valid for Monte Carlo rendering.
In contrast, our neural reflectance field representation models \emph{all camera-light interactions with the scene}. In addition, it is trained in conjunction with a physically-based ray marching framework.
As a result, it can be easily integrated using standard graphics rendering engines, by simply implementing the reflectance function as a special phase function.

In particular, we use Mitsuba \cite{mitsuba} to render one of our captured \nrfs{} under complex environment illumination, and show these results in Fig.~\ref{fig:mituba}.
We simply compute discrete $512\times 512\times 512$ volumes from our reflectance fields and use the volume to do Monte Carlo rendering.
While simple, this leads to very realistic rendering results in Fig.~\ref{fig:mituba}.
Also note that this allows us to compose a scene that is made up of our captured object and traditional 3D models represented by meshes with BRDFs, and \emph{simulate the light transport between these different representations} including complex shadows and inter-reflections. 
While these results contain fewer details compared to our other results, this is caused by the limited volume resolution and can be addressed by potentially implementing our network in Mitsuba. 


\paragraph{Limitations.} 
Our method is able to produce high-frequency appearance effects with 
fine details in most cases. However, it may still result in slightly blurry results when there are too many details (like the results in Fig.~\ref{fig:fur} and \ref{fig:face}). Increasing the network capacity could potentially  
alleviate this. While our method generally generates a clean background without requiring any masks, some minor dark floaters occasionally appear,
mainly coming from background regions that are not dark enough and are seen by several views. This usually can be addressed by masking the volume density in 3D with a bounding box.
Our adaptive transmittance is efficient, but it may introduce some minor flickering in videos when doing relighting, due to inconsistent adaptive samples across frames. Increasing the number of samples in the volume usually resolves this. 
Some of these issues are visible in the supplementary video.

\section{Conclusion}
We present a deep learning based approach for appearance acquisition using a simple mobile phone setup.
We present a novel neural reflectance field representation, which encodes volume rendering properties to model the geometry and reflectance of real scenes.
We leverage a differentiable physically based ray marching framework to learn the \nrf{} in a scene-dependent deep training process.
We demonstrate that our \nrf{} can be effectively estimated from cellphone flash images under collocated camera and light, 
allowing us to render photo-realistic images under arbitrary camera and (non-collocated) light positions.
Our method is able to generate high-quality relighting and view synthesis results, reproducing challenging appearance effects, such as specularities, shadows, occlusions, and fine textures, which are significantly better than results from previous mesh-based and volume-based methods.
Moreover, since our \nrf{} are learned in a physically based rendering framework, they can be also rendered in standard graphics rendering engines, enabling scene modeling applications.
Our approach takes a step towards making neural capture and rendering more practical and compatible with standard graphics pipelines.

\Comment{
\begin{figure}[t]
    \vspace{1in}
    \caption{Material editing.}
    \label{fig:material}
\end{figure}
}

\section*{Acknowledgements}
This work was supported in part by ONR grants N000141712687, N000141912293, N000142012529, NSF grant 1617234, Adobe, the Ronald L. Graham Chair and the UC San Diego Center for Visual Computing.

\bibliographystyle{ACM-Reference-Format}
\bibliography{reference}

\end{document}